\documentclass[conference]{IEEEtran}
\IEEEoverridecommandlockouts
\usepackage{cite}
\usepackage{amsmath,amssymb,amsfonts}
\usepackage{algorithmic}
\usepackage{graphicx}
\usepackage{textcomp}
\usepackage{xcolor}

\usepackage{multirow} 
\usepackage{multicol} 

\def\BibTeX{{\rm B\kern-.05em{\sc i\kern-.025em b}\kern-.08em
    T\kern-.1667em\lower.7ex\hbox{E}\kern-.125emX}}
\begin{document}

\title{Intrinsic Image Captioning Evaluation \\
{\footnotesize \textsuperscript{}
}  
\thanks{
} 
}



\author{\IEEEauthorblockN{Chao Zeng}
	\IEEEauthorblockA{\textit{City University of Hong Kong} \\
		\textit{chao.zeng@my.cityu.edu.hk}\\
		}
	\and
	\IEEEauthorblockN{Sam Kwong}
	\IEEEauthorblockA{\textit{City University of Hong Kong} \\
		\textit{cssamk@cityu.edu.hk}\\
		}

}

\maketitle

\begin{abstract}
The image captioning task is about to generate suitable descriptions from images. For this task there can be several challenges such as accuracy, fluency and diversity. However there are few metrics that can cover all these properties while evaluating results of captioning models.In this paper we first conduct a comprehensive investigation on contemporary metrics. Motivated by the auto-encoder mechanism and the research advances of word embeddings we propose a learning based metrics for image captioning, which we call Intrinsic Image Captioning evaluation ($I^2CE$). We select several state-of-the-art image captioning models and test their performances on MS COCO dataset with respects to both contemporary metrics and the proposed $I^2CE$. Experiment results show that our proposed method can keep robust performance and give more flexible scores to candidate captions when encountered with semantic similar expression or less aligned semantics. On this concern the proposed metric could serve as a novel indicator on the intrinsic information between captions, which may be complementary to the existing ones.
\end{abstract}

\begin{IEEEkeywords}
image captioning, auto-encoder, evaluation metric, word embeddings, sentence similarity
\end{IEEEkeywords}

\section{Introduction}
In recent years there has been abundant images uploaded in the Internet. Generating proper descriptions from visual images to language expression has gained much attention from computer vision and natural language processing researchers. And since neural networks prosper the research community has witnessed proliferation of various neural image captioning models. However The development of evaluation metrics for image captioning seem to remain unchanged for years, such as BLEU, METEOR, ROUGE or CIDER\cite{b1}\cite{b2}\cite{b3}\cite{b4}, all of which are about to calculate a hard alignment between candidate captions and ground truth sentences. More recently the SPICE metric use scene graph to add the visual information when comparing the similarity between candidates and ground truth and shows cross-modal similarity\cite{b12}.

An interesting phenomenon emerged in recent years on MSCOCO Image Captioning Challenge is that the neural models are now outperforming human performance with respects to the contemporary evaluation metrics, which we may consider not convincing and need to give deeper investigation towards the automatic metrics\cite{b8}. 

Generally these kinds of metrics provide with different approaches to calculating the similarity degree between the generated captions and the ground truth captions in the dataset. Some of these contemporary metrics are actually from related research area. For example, the BLEU metric is from machine translation. The ROUGE metric is designed for text summary originally. METEOR is another metric also borrowed from machine translation task. CIDER and SPICE are specially designed for image captioning, which are derived in more recent research studies
.
There are some key challenges faced with the automatic evaluation metrics for image captioning. Firstly the contemporary metrics are struggling with the problem of deviating from human judgements with its nearly mechanical manner of comparing on candidate captions\cite{b5}. Thus metrics based on token overlap would miss the semantic similarity among sentences. For one thing there can be various expression pointing to the same meaning, for another there might be altering meanings of the same word, which are all difficult to cover for the traditional metrics. Secondly there can be various blind spot for rule based metrics. For example, the SPICE metric excels in capturing the visual information by scene graphs but not as good when it comes to sentence structural information in sentence level.

To address the above mentioned challenges and inspired from auto-encoding mechanism, we propose a learning based metric that captures intrinsic information entailed among different sentences. The main idea of auto-encoder is to recover the input from itself. To be more specific in an auto-encoder the input would be first transformed into a vector representation with an encoder, followed by a decoder to generate the original input. Here the supervision information is from the input data itself. Inspired from this and we assume that if the original data can be recovered by the intrinsic representation then the similarity between the representation should be an indicator as the similarity of the original data. Based on the above idea we carry out the proposed Intrinsic Image Captioning Evaluation ($I^2CE$). We use pre-trained GloVe vectors as word representation and encode the input sentence with GRU. Similarly at the decoder side we use GRU to decode the encoded information together with self attention mechanism. We combine some translation corpus and MSCOCO sentences to train the sentence auto-encoder. And we select several state-of-the-art image captioning models to test their performances with respects of both conventional metrics and our $I^2CE$. Experiments results show the effectiveness and robustness towards captions of different qualities.
The main contributions of this paper include:

\textbf{·} We propose the Intrinsic Image Captioning Metric ($I^2CE$), a learning based metrics based on auto-encoder mechanism and word embeddings.

\textbf{·} We demonstrate how to train an auto-encoder with sentence corpus and serve as captioning evaluator.

\textbf{·} We test performances of various state-of-the-art image captioning models on the MSCOCO dataset with both contemporary metrics and our proposed method. Our metric show dynamic and highly semantic related properties on scoring for testing captions.

\section{Related Work}

\subsection{Contemporary Adopted Metrics}

\[{\rm{BLEU}} = \frac{{\sum\nolimits_i {\sum\nolimits_k {\min \{ {h_k}({c_i}),\mathop {\max }\limits_j h({s_{ij}})\} } } }}{{\sum\nolimits_i {\sum\nolimits_k {{h_k}({c_i})} } }}\]

$BLEU$ is originally designed for machine translation evaluation\cite{1}. In the equation of the definition, $k$ and $i$ are the index of the N-gram token and the testing case. The $c$ and $s$ stands for candidate and reference caption respectively. The $h$ is the number of occurrences of the token. So $h_k(c_i)$ means the ${k}_{th}$ word token’s number of occurrences in the $i_{th}$ candidate caption. From the original metric expression formula we can see that this is actually describing the proportion that how much of the generated N-grams got matched with the ground truth caption references out of all the N-grams of the candidate sentence. To give the matching less strict it takes the maximum number of the occurrences accounting among comparing with all the references of the dataset(generally there are 5 ground truth references for each image). To avoid single word repeating style of candidate scoring full mark it takes then the minimum between the numbers of occurrences in candidate and in ground truth. After iterating all the N-grams in the candidate caption then the contributions of different N-grams(N equals to 1,2,3, or 4) are then summed to get the final score of the candidate sentence with the short length punishing term attached. Eventually we get the BLEU scores.

$ROUGE$. The ROUGE metric \cite{b3} is very similar to BLEU metric but gives specifically concerns on the balance between the recall and precision, in which case the BLEU only consider the precision of the generated candidate while the recall not covered. And this kind of metric was originally proposed for text summary tasks.

$METEOR$. $METEOR$ is also designed for machine translation\cite{b2}. Again this may be considered as another enhanced version of original BLEU matric. Based on the F measure of the previous ROUGE metric, here the novel ingredient is the introducing modifier of chunk punishment term. Automatically this algorithm will find the non-intersecting matching chunk pairs between the candidate and the reference. If the candidate strictly matches with the reference then this term reaches its minimum and in this case the METEOR score is at its maximum level. 

$CIDER$. The core mechanism embedded in CIDER is the tf-idf weighting term\cite{b4}, which has wide range of application in vector semantics. The main idea of this term comes from the assumption that the occurrence frequency of a word in a certain corpus can be a significant indicator for the character or semantic meaning of the word. Some words occurs frequently in nearly every corpus(“and”,”the”,etc.). So when we consider the meaning of a whole sentence we need to weight different words for significance on this concern.

$SPICE$. Instead of considering matching with N-grams in the caption like previous metrics, SPICE creatively employ word tuples and calculate the intersection between candidate and ground truth captions. Another highlight of SPICE metric is the graph representation of sentences.
The SPICE metric\cite{b12} covers more information than N-gram token. The matching process is carried out through the comparison between two tuple sets of texts and image graphs. However the performance of this metric relies much on the parsing accuracy. It prefers longer sentences and tend to lose structural information of sentences\cite{b5}.

\subsection{More Recently Proposed Metrics}
Learning based metrics. This kinds of metrics are different from the traditional rule based ones. Cui Yin et al. proposed a binary classifier as the caption metric model\cite{b5}. The model are trained by two kinds of captions: the human drafted ones and the machine generated ones which are labelled one or zero. Finally the distinguishing model would give a score which is between zero and one, indicating how likely a caption is generated by human. And the authors take this probability score as a metric for image caption on the concern of human relativeness.

Retrieval model. Ruotian Luo et al. introduce an image retrieval auxiliary image caption generation model\cite{b10}. In the experiment the authors mainly explore with the diversity oriented loss terms for generating more diverse captions. For measuring the singularity of the generated captions they utilize an image retrieval model to provide with feedbacks as an indicator and relay this extra signal back to the caption generation model.  The advantage a generated caption may have in image retrieval process could be another kind of metric for the quality of a candidate caption.

WMD metric takes the similarity problem between candidate and reference captions as a special instance of the Earth Mover’s Distance, which is a well known transportation linear optimization problem\cite{b6}.

\[\mathop {\min }\limits_{T \ge 0} \sum\limits_{i,j}^n {{T_{ij}}c(i,j)} \]

In this optimization objective the $c_{ij}$ is the Euclidian distance from word $w_i$ to word $w_j$ in semantic space. The $T_{ij}$ is the transport amount between word $w_i$ to the warehouse word $w_j$. The amount is the weight of occurrence  a word carries in the whole sentence. The assumption is that the more similar the two captions the less cost it would take to transfer one to the other.

Diversity metric. On the concern on diversity in image caption generation Qinzhong Wang et al. proposed a diversity indicator as a metric\cite{b10}. The main idea is that for a specific generation model and a specific image let it generate for say three different captions and then conduct a k-SVD decomposition analysis on these captions. If the singularity values are balanced well then we can say the model is good at diversity property. If not--the singularity value is biased to a single value and the rest are all near zero, then we know the tree captions are actually of the same meaning, in another word the model is less versatile. 

\subsection{Discussion on Existing Metrics}
From the above discussion we can see that actually BLEU, ROUGE, METEOR and SPICE are all based on  matching of word tokens with various styles. For the original BLEU the contributions of the score consists of discrete binary terms. In this case we only add the contribution when the current considered token are strictly matched or the same with that of the ground truth caption. When encountered with synonym or semantic similar tokens their contribution would be missed in the numerator term.

The SPICE metric uses tuple sets in stead of N-grams for comparing. To ameliorate the performance on synonym encountering some of them utilize the WordNet as a soft mechanism to enhance the original matching process. As we have discussed previously the hard matching manner proposed by this style of metrics would let us miss the contributions of those semantic diverse but similar expressions in the candidate sentences. 

The more recent ones are constrained by different practical concerns. For example, the WMD metric is also a soft matching method, but originally this distance is for document and may not be suitable for shorter sentences. This work is the most similar one to us among all the above metrics. But the main mechanism behind is for measuring document distance while the caption sentences are usually much shorter relatively. And thus it may not necessary to use the occurrence weights for different words in the caption since nearly each word just appears for once especially for the key words. This can be indicated by the huge sparsity of the transport matrix in the metric. 

Motivated by above mentioned concerns we introduce a soft matching metric called Intrinsic Image Captioning Metric to encourage more diverse expressions towards the considered image, which may actually keep original intrinsic semantic information and are just phrased in different manners.

\section{The proposed Method}
Our inspiration comes from vector semantics. As once a philosopher put it, “the meaning of a word is its usage in the language”(Wittgenstein,1953). And thus we can use the concurrences among different words in abundant amounts of corpus to represent the sense of a single word. This representation is called vector semantic. And with this method now we can map word which is originally string of characters to a vector in semantic space. We are inspired by this idea and use pre-trained GloVe vectors and then employ an auto-encoder to form the sentence level representation. With the above idea we  introduce our Intrinsic Image Captioning Evaluation.

\begin{figure}[htbp]
\centerline{\includegraphics[width=9.5cm]{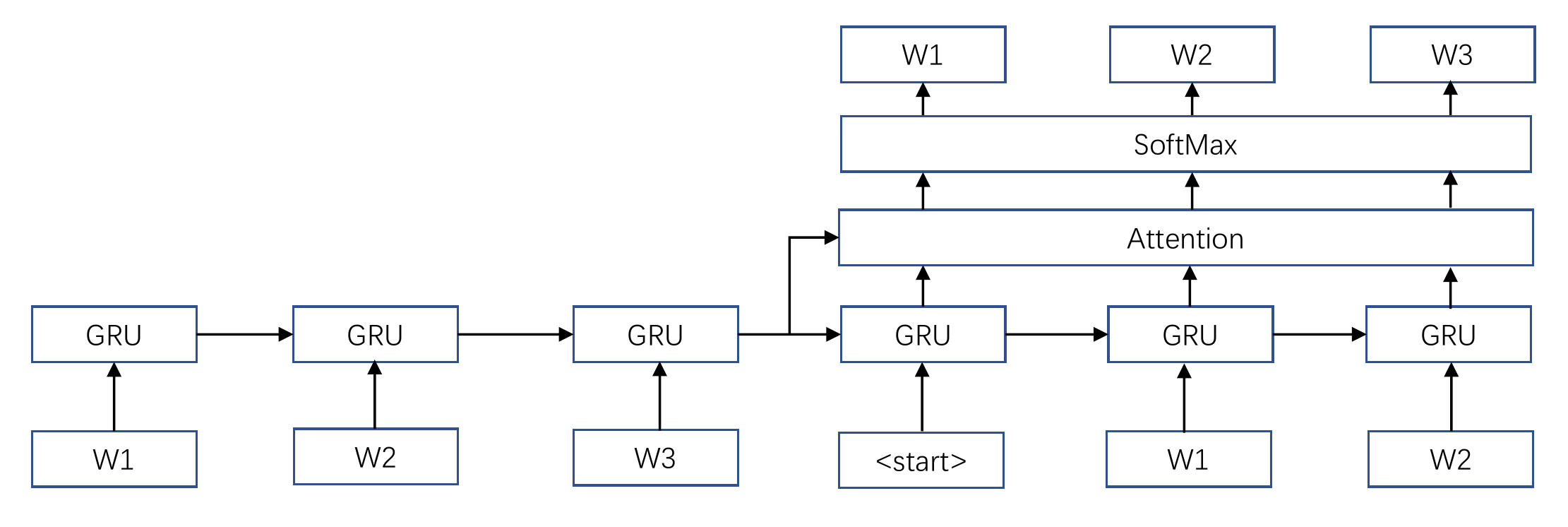}}
\caption{$I^2CE$ system framework.}
\label{fig}
\end{figure}

As shown in the above Fig. 1, we employ an auto-encoder for the word vector fusion. The left half is the encoder for extracting the meaning of the sentence, which is represented by the last hidden state vector of the encoder, which we name Intrinsic Vector. The right half is the decoder. It is only employed in the training stage to make learning the intrinsic encoder possible. 

The assumption here is that if the auto-encoder can reconstruct the input sentence itself after abundant training process then we can say the intrinsic vector between the encoder and decoder can be taken as the meaning of the input sentence. In order to take advantage of Word2vec  representation and the large corpus training datasets we use pre-trained GloVe vectors to initialize the embedding layer for the encoder. In addition the Bahdanau attention is utilized for the decoder for better performance\cite{b33}. And the decoder is built just like traditional RNN based sequence decoder. Since we use the input sentence itself as supervision for the learning process our model is learnt in a self-supervised manner.

After abundant training process we utilize the encoder to conduct sentence representation and get the intrinsic vector for calculating the similarity between the candidate and reference captions.

\subsection{Preprocessing on Sentences}\label{AA}
To illustrate the difference between hard and soft matching we take consideration on the stop words in caption. And for this concern we propose a simpler version of soft matching metric $I^2CE$.
As we want to utilize the mapping between the word and vector we need first to extract words from the candidate captions. For the $MEAN$ metric, we consider the meaning of a sentence represented by its core words. And the words that are less important(which can be called stop words)for understanding the whole sentence would be removed before evaluation. We adopt the standard definition set of stop word provided by $NLTK$ library.

On the other hand for our proposed $I^2CE$ metric we need to map the words in caption into vectors with the pre-trained GloVe vectors. Also we need to build the index to word mapping so that we get the word index to vector mapping matrix. With this matrix we initialize the embedding layer in encoder as constant and fix it in the training process of the auto-encoder.

\subsection{Word Vector Learning}
Originally a word vector were generated by counting the number of times it occurs in different corpus. Then the words co-occurrence matrix was introduced instead. In this way we may get a word represented by a long dimensional and sparse vector. However, the short and dense vector based word representation models like GloVe, Fasttext have shown great performance on various NLP tasks. 
The learning mechanism behind is the assumption that words of similar meaning appear neighboring with each other in language corpus. Or we may put it in another way the meaning of a word is defined by its usage.
 
The Word2vec algorithm\cite{b7} provides with a good strategy for learning dense word vector representations. The main idea is that instead of counting now we consider a binary classification problem on how likely a word occurs as part of the context of a target word. With this idea we can simply use two separate linear layers to map the original one hot vectors of target and context words to dense vectors. Then we compare the mapping results and carry out the binary classification process through a soft-max layer. Finally at the output side of soft-max we need to compare the output results and the ground truth about word co-occurrence and calculate the difference for back propagation. The optimization objective is two folds. First to maximize the co-occurrence of true (target, context) pairs and minimize negative pairs. After abundant numbers of  iterations the mapping layer would capture the principles of word co-occurrences and can generate proper dense vectors.

\[{\rm{L}}(\theta ){\rm{ = }}\log \frac{1}{{1 + {e^{ - ct}}}} + \sum\limits_{i = 1}^k {\log \frac{1}{{1 + {e^{{n_i} \cdot t}}}}} \]

Above is the learning objective to maximize for the training process. There are generally two different strategies in word2vec algorithm, the continuous bag of words and the skip-gram model. Here the GloVe vector we select adopts the skip-gram. In this formula the $c$, $t$ stand for context and target word vector respectively. And $n_i$ is the noise word by the $NCE$ training trick\cite{99}. We can see that the objective is actually both maximizing the probabilities of the true contexts and minimizing that of the noisy words.

\subsection{Word Vector Fusion with Auto-Encoder}
After we map the words into the semantic space. Then the meaning of a sentence can be considered as a cluster of vectors or points in the representation space. 
\[MEAN = \frac{{\sum\limits_{i = 1}^k {{w_i}} }}{k}\]
Basically to introduce the vector semantic idea to the sentence similarity we may make simplified the representation of the meaning of a sentence by taking the centroid of the corresponding sentence cluster, which is shown as the above equation. The $w_i$ is the vector representation of the $i_th$ word and $k$ is the number of words in the sentence. This can be a metric for image caption we call $MEAN$, a simplified version of soft matching metric based on vector semantics.

To take a step further and enable the metric to capture more significant features on a candidate sentence we utilize a GRU based auto-encoder to do the fusion on word vectors, involved with the whole candidate caption. This metric is our proposed $I^2CE$.

Auto-encoder is a forward neural network generally with a pair of encoder and decoder. Or we may put it in another way the auto-encoder is actually a data compression algorithm, which is of data driven and lossy. However, in contrast to traditional rule based encoder and decoder framework the auto-encoder is built with neural network generally. And there may be some common properties shared by different versions of auto-encoders. Firstly, they are data driven models and large datasets are needed for the training process. Since they are data driven and thus they may have data bias inherently and the domain adaptation is a great challenge for this concern. Secondly, the auto-encoder is a lossy compression model and the reconstructed data may be degraded in some scale. In addition the auto-encoder is a self supervised learning algorithm without any other label data, which makes it  more convenient for training.

In the case of our paper we built the encoder and decoder with the Gated Recurrent Units. For the encoder to utilize the advantages of large corpus we initialize the embedding layer as constant and keep it fixed in the training process of the auto-encoder. And for the decoder the Bahdanau Attention is employed, which was originally introduced for machine translation\cite{b33}. To be more specific we use the GRU to encode the input sequence of sentence. For training the encoder we need a paired decoder with the self supervision. After training with abundant corpus we take the encoder as our core module to calculate the intrinsic vector of a specific sentence.

\subsection{Cosine Similarity for Soft Matching}
To eventually give the similarity score between the candidate and ground truth caption we need a distance metric for comparing the previous defined Intrinsic vector. Up until now the widely adopted one is cosine similarity. 

\[Sim(c,{s_j}) = \frac{{I(c) \cdot I({s_j})}}{{|I(c)| \cdot |I({s_j})|}}\]

In the above equation $I$ is the intrinsic vector of the sentence. The $c$ and $s$ are the candidate and reference caption respectively. This can be understood as that the more likely two captions describing similar meaning the more likely their semantic vector appear in neighbor area in the semantic space. For the $MEAN$ method we can simply use the averaging embedding of core word vectors as the intrinsic vector of the corresponding . And by averaging the $MEAN$ scores between candidate and all the references we would get the final evaluation score for a single sentence. And by taking the mean of all candidate cases in the testing dataset we get the final $MEAN$ evaluation score for the testing model.

For our proposed $I^2CE$, the intrinsic vector I in equation should be the intrinsic vector extracted by the metric encoder. And by computing the similarity between the intrinsic vectors of (candidate, reference) pair we got the score of the candidate giving regards to the considered ground truth sentence.

\section{Experimental Setup}
The main objective of this part is to validate the effectiveness of the proposed metric. As discussed in previous sections, the popular metrics like $BLEU$ try to find matching tokens in the reference captions which depends largely on the phrasing style in the process of caption generation.

We first introduce the dataset employed to test the performances of both hard matching and soft matching. Then we briefly list the adopted image caption generation models.

$Dataset$. Firstly for the training of the caption models. Based on the Microsoft Common Objects in context(MSCOCO) dataset, Microsoft research team introduced the MSCOCO Caption dataset. This dataset has two branches. One is MSCOCO-C5, In this branch the training, validation and testing splits have the same image collection as the original COCO dataset. With each image there are five manual label sentences as reference captions.
The other is MSCOCO-C40. Each image has forty labelling sentences in this branch of dataset. Since the labelling work is very time consuming there are only five thousand images in total for this branch. We choose the MSCOCO-C5 branch as our evaluation dataset for better generalization.

Secondly, For the training of the proposed metric model $I^2CE$. We first train a rudiment sentence encoder with the Spanish to English Translation Dataset. The training trick here is that only use the English half of the original corpus, which contains up to one hundred thousand sentences. So we input a English sentence and expect to output itself in the decoder side. After that we then adapt the auto-encoder to the captioning corpus, the annotation or reference sentences in the MSCOCO-C5 dataset.

$Image Caption Models$. In this paper we employ some of the current popular generation models including show and tell\cite{b15}, top-down attention\cite{b9}, attention2in\cite{b26}, top-down with self-critical training and attention2in with self-critical training.

Show and Tell. The Show and Tell is one of the earliest image caption models based on neural network. The main idea is two steps for caption generation. First to use convolution neural network to extract image feature. Second to use the recurrent neural network to modelling the language generation process, in which the final sentence is generated word by word and the previous generated ones are taken as the input for next time step.

Top-down attention. The Top-down Attention model enhances the original neural models by various enhancements. They use the fast R-CNN to analysis the image first to extract some attributes. Guiding by this pre-generated attributes the model can generate caption in a more accurate manner, which aligns more with the visual information. Since this top-down manner the model is then called top-down attention.

Att2in. The Att2in is similar to fully connected neural model enhanced by adding attention term to LSTM cell node. The proposers find that when optimizing with Adam optimized this structural altering leads to better performance.

Self-Critical Sequence Training. The SCST training\cite{b26} utilizes reinforcement learning as training procedure for the caption generator. In this case the agent is the LSTM model within the environment of visual and text features. And the action is about which word to choose at each time step. However, different form traditional modelling of reinforcement problem, the method use the generated sentences on testing stage as baseline instead of directly calculating the reward. And the better ones than the baseline would be weighted positively or negatively if not.

\section{Experimental Results}
In this section we first compare the performance of the above mentioned image generation models with regard to the contemporary caption metrics--BLEU, METEOR, ROUGE, CIDER and SPICE. A hard matching mechanism is embedded in these rule based metrics. We simply choose the BLEU metric to modify and form our proposed $I^2CE$ metric. Then we compare the metric scores of the two different matching strategies. And Finally we include evaluation scores of captions calculated by $I^2CE$ and compare all previous results together. In addition, we also give some intuitive $I^2CE$ scores of both high quality generated captions and also the captions less aligned to reference ground truth.

\subsection{Soft Matching vs. Hard Matching}
As shown in Table 1, when the non-significant stop words are removed we compare the scoring results of the comparing models with regards to both the hard matching and the $MEAN$ metric. The left column is the scores of original hard matching $BLEU$. We can see that after removing the stop words the $BLEU$ score decrease in a significant scale. 

\begin{table}[htbp]
    
	\centering  %
	\caption{Comparing Bleu@1 and the proposed metric when the stop words removed} 
	\label{table1}
	
	
	\begin{tabular}{|c|c|c|c|c|c|}  
		\hline  
		\textbf{Caption}&\multicolumn{5}{|c|}{\textbf{Captioning Models}} \\
\cline{2-6} 
		& & & & &\\[-6pt]  %
		
		\textbf{Metric}&show-tell&top-down&atten2in&top-down+&atten2in+ \\  
		
		\hline
		& & & & &\\[-6pt]  %
		B@1&0.700&0.749&0.735&0.779&0.767 \\
		
		\hline
		& & & & &\\[-6pt]  %
		B@1*&0.376&0.418&0.418&0.488&0.493 \\ 
		
		\hline  
		& & & & &\\[-6pt]  
		GloVe&0.803&	0.814&	0.805&	0.826&	0.815 \\ 
		
		\hline
	\end{tabular}
\end{table}

For this we think the reason is that in the original $BLEU$ metric the stop words make a considerable contribution towards the total nominator term, which we think is less informative for the semantic representation of a sentence.
In addition we notice that on $BlEU@1$ metric the $att2in+$ outperforms the top-down+ while the inverse is true for the soft matching $MEAN$ metric. The reason for this phenomenon may be that for candidate captions generated by $att2in+$ there are relatively more stop words matching contribution while for the $top-down+$ more matching contributions are made by key words.

\subsection{Comparing with Contemporary Metrics}
The Table 2 compares our soft matching based metrics $MEAN$ and $I^2CE$ with the popular metrics on state-of-the-art captioning models. From the above results we can see generally for better models recognized by popular metrics such as $top-down+$ and $att2in+$ our $MEAN$ and $I^2CE$ also would give relative higher scores for them. And comparing to the $MEAN$ which simply take averaging as fusion method, the $I^2CE$ seems to be more distinguishable since the scores distribute in relative larger range. This can be taken as that the averaging fusion may have the side effect of blurring on the representation of the whole sentence composed of its key words.

\begin{table}[htbp]
	\centering  
	\caption{Performance Evaluation on $I^2CE$ metric and contemporary automatic metrics}  %
	\label{table}  %
	
	\begin{tabular}{|c|c|c|c|c|c|}  
		\hline  
		
		\textbf{Caption}&\multicolumn{5}{|c|}{\textbf{Captioning Models}} \\
\cline{2-6} 
		& & & & &\\[-6pt]  %
		
		\textbf{Metric}&show-tell&top-down&atten2in&top-down+&atten2in+ \\  
		
		\hline
		& & & & &\\[-6pt]  
		B@1&0.700&0.749&0.735&0.779&0.767 \\
		
		\hline
		& & & & &\\[-6pt]  %
		B@2&0.528&0.582&0.567&0.612&0.603 \\
		
		\hline  
		& & & & &\\[-6pt]  %
		B@3&0.387&	0.437&	0.423&	0.459&	0.451 \\  %
		
		\hline  
		& & & & &\\[-6pt]  %
		B@4&0.282&	0.323&	0.312&	0.338&	0.333 \\  
		
		\hline  
		& & & & &\\[-6pt]  %
		Meteor&0.234&	0.259&	0.250&	0.263&	0.257 \\  
		
		\hline  
		& & & & &\\[-6pt]  %
		Rouge&0.514&	0.548&	0.536&	0.556&	0.552 \\  
		
		\hline  
		& & & & &\\[-6pt]  %
		CIDEr&0.901&	1.065&	1.019&	1.182&	1.156 \\ 
		
		\hline  
		& & & & &\\[-6pt]  %
		SPICE&0.164&	0.191&	0.184&	0.195&	0.189 \\ 
		
		\hline  
		& & & & &\\[-6pt]  %
		\textbf{$I^2CE$}&0.751&	0.731&	0.732&	0.789&	0.822 \\  
		
		\hline
	\end{tabular}
\end{table}

With the knowledge learned from large language corpus the utilized auto-encoder would have a better ability on distinguishing caption semantics. In addition $MEAN$ removes the stop words for both candidate and reference sentences which may cause more loosen tendency on the evaluation scores. That is to say all the comparing models are capable of capturing the key objects or attributes in the scene but with various level of expressions, in which stage the stop words may take significant effects. In contrast $I^2CE$ consider on the original whole sentences and extract semantics with learnt prior. From the results comparing to $MEAN$ we can see that it is better at distinguishing different caption models with various levels of expression since it has a larger interval of scores distribution. In addition, comparing with the original $BLEU$ the scores are higher since it employs the soft matching strategy.

\subsection{Intuitive Results on $I^2CE$ Scores}
The following Fig.1 shows us the scores given by $I^2CE$ on some testing cases with the testing captioning model $att2in+$. In the figure there are three pieces of testing cases shown. For each testing case there are one candidate caption in the first line and the five references below. Addicted to each reference is the corresponding $I^2CE$ score for the (candidate, reference) pair.

\begin{figure}[htbp]
\centerline{\includegraphics[width=9.5cm]{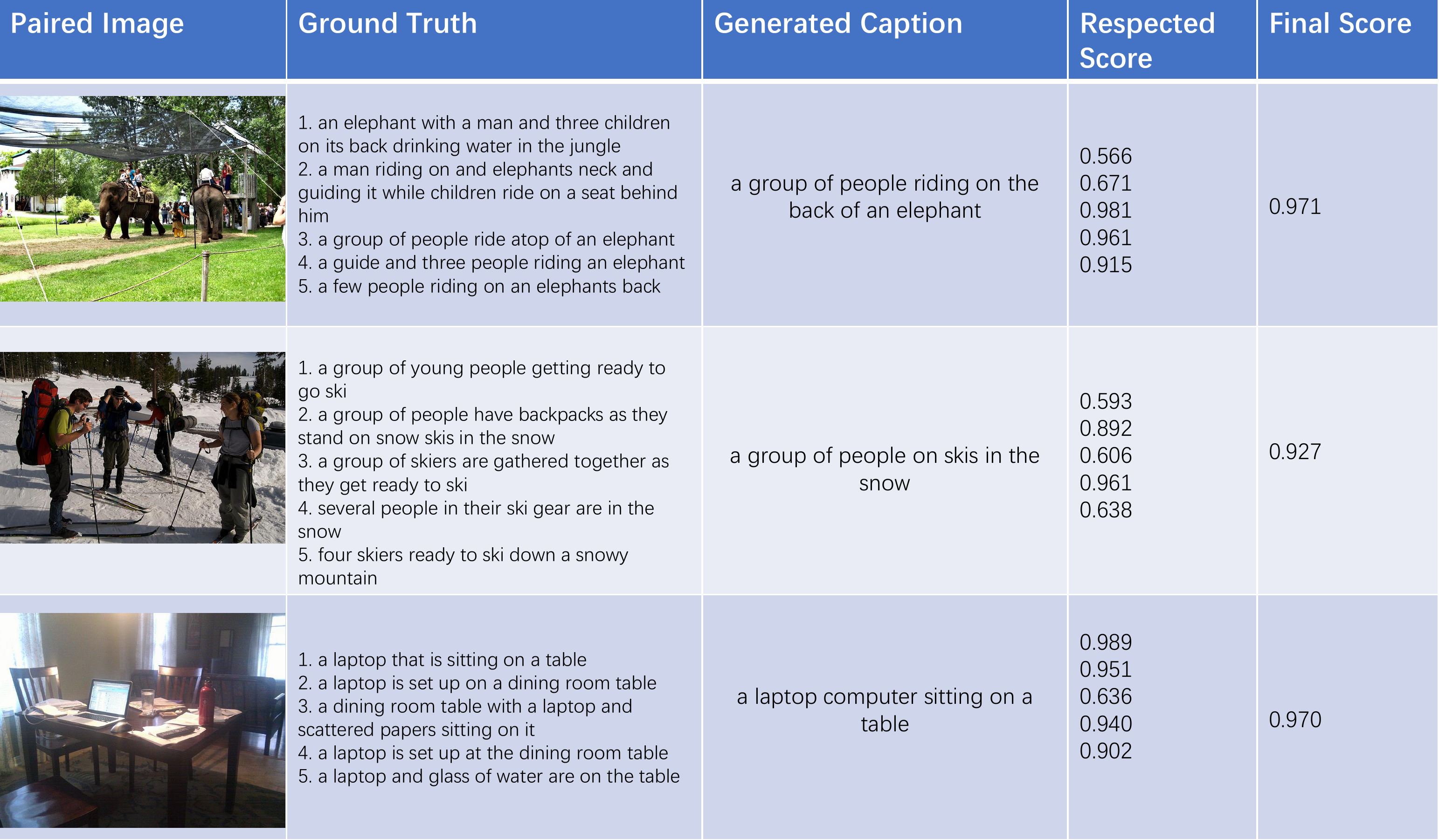}}
\caption{$I^2CE$ scores on highly aligned generated captions.}
\label{fig}
\end{figure}

Take the first picture for example, we can see that for the latter three generated captions which are most similar references given the candidate caption get relatively higher scores. For the first reference it has some additional details like ``drinking water" and ``in the jungle", which are not covered by the candidate caption and thus the similarity score is lower. The second reference is quite likely the same case. However the latter three references are just expressing exactly the same information as the test caption. Thus the candidate would reach higher scores.

\begin{figure}[htbp]
\centerline{\includegraphics[width=9.5cm]{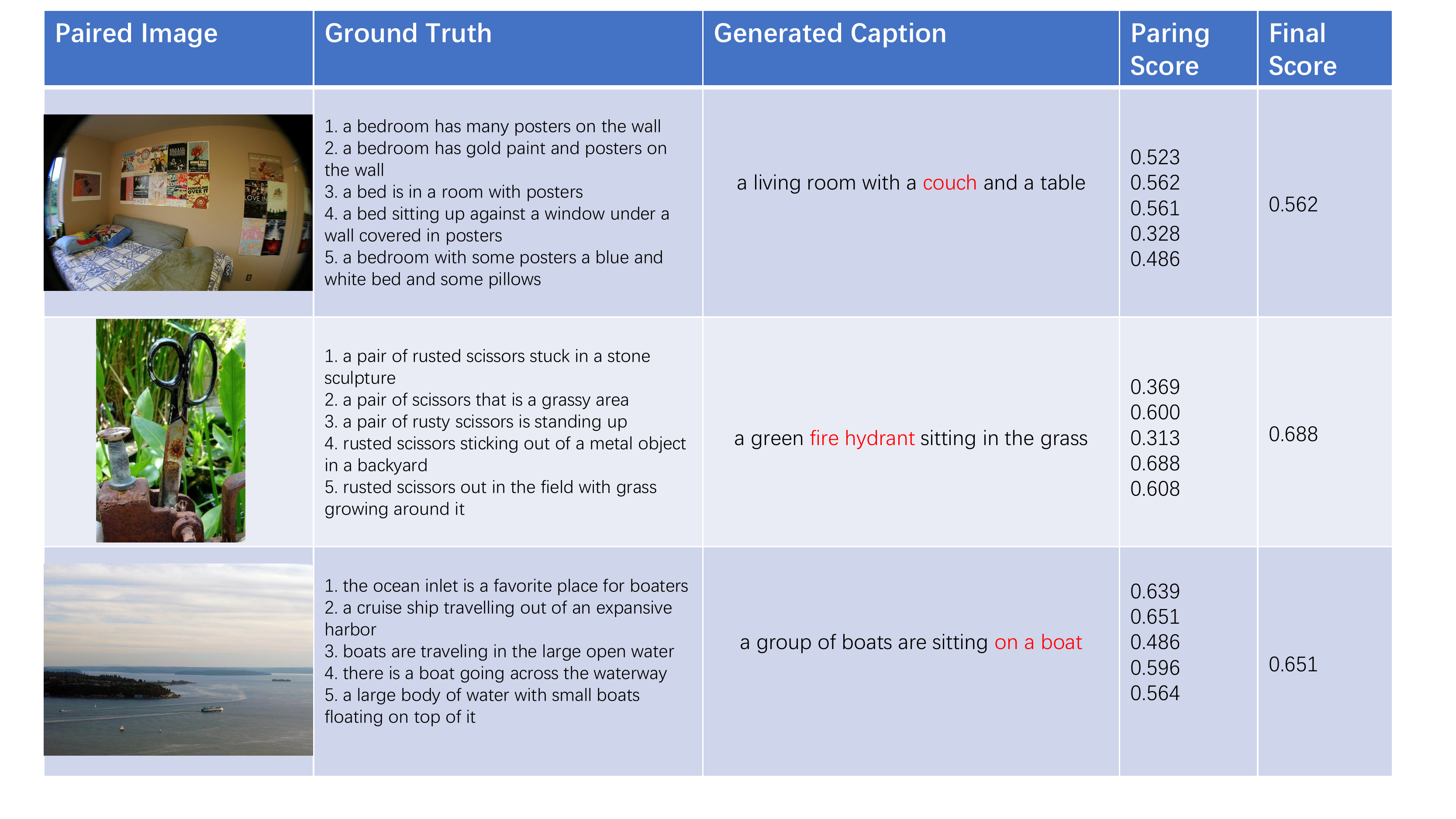}}
\caption{$I^2CE$ scores on less aligned generated captions.}
\label{fig}
\end{figure}

To observe the scoring results in a larger scope  we present some lower scored cases given in Fig.2. As we can see, for these captions less aligned to the references there are whether miss detected conceptions like ``couch" and ``fire hydrant", or pathological syntactic sentence structure.

\section{Conclusion}
In this paper, we give special concern on the matching strategy in image caption evaluation metrics. We first take a comprehensive investigation and in depth analysis on the core mechanism behind the contemporary metrics including $BLEU$, $ROUGE$, $METEOR$, $CIDER$ and $SPICE$. In addition, we also give respects towards the more recent proposed image caption metrics like the learning based, $Self-CIDER$ and $WMD$ distance. Then we introduce our proposed $I^2CE$ metric, which use the sentence intrinsic vectors to calculate similarity instead of hard matching on word tokens. To keep the syntax structure of the original sentence we take a step further and employ an auto-encoder to conduct vector fusion on word representation.

To validate the effectiveness of the proposed soft matching method we conduct three stage experiments. First we evaluate the performances of state of art models with the contemporary metrics. And then we analysis the results giving regards to the latent relationship between these different metrics. Further in the next stage we remove the contribution of stop words and take observation on the matching performance of the comparing models. And Finally we compare the scoring results of our proposed $I^2CE$ metric with all the previous metrics. Experiment results show the proposed method can give more reasonable and flexible evaluation scores when encountering synonymous expressions. Future work may relate to better word vector fusion methods and perhaps pre-trained based sentence vector learning strategies for image caption evaluation.

\section*{Acknowledgment}

The authors would like to thank the anonymous reviewers
for their valuable suggestions. This work is supported
by UGC of HKSAR.


\begin{thebibliography}{99}
\bibitem{b1} Kishore Papineni, Salim Roukos, Todd Ward, and Wei-Jing Zhu. 2002. BLEU: a Method for Automatic Evaluation of Machine Translation. Proceedings of the 40th Annual Meeting of the Association for Computational Linguistics (ACL), Philadelphia, July 2002, pp. 311-318
\bibitem{b2} Satanjeev Banerjee and Alon Lavie. 2005. METEOR: An Automatic Metric for MT Evaluation with Improved Correlation with Human Judgments. Proceedings of the ACL Workshop on Intrinsic and Extrinsic Evaluation Measures for Machine Translation and Summarization
\bibitem{b3} Chin-Yew Lin. 2004. ROUGE: A Package for Automatic Evaluation of Summaries. Proceedings of the Annual Meeting of the Association for Computational Linguistics (ACL).

\bibitem{b4} Ramakrishna Vedantam, C. Lawrence Zitnick, and Devi Parikh. 2015. CIDEr: Consensus-based Image Description Evaluation. In CVPR.
\bibitem{b5} Yin Cui, Guandao Yang, Andreas Veit, Xun Huang and Serge Belongie. 2018. Learning to Evaluate Image Captioning. In CVPR.
\bibitem{b6} Mert Kilickaya, Aykut Erdem, Nazli Ikizler-Cinbis, and Erkut Erdem. 2018. Re-evaluating Automatic Metrics for Image Captioning. In ACL.
\bibitem{b7} Quoc Le, and Tomas Mikolov. 2014. Distributed Representations of Sentences and Documents. In Proceedings of the 31 st International Conference on Machine Learning, Beijing, China.
\bibitem{b8} Qingzhong Wang and Antoni B. Chan. 2019. Describing like Humans: on Diversity in Image Captioning. In CVPR.
\bibitem{b9} Peter Anderson, Xiaodong He, Chris Buehler and Damien Teney. 2017. Bottom-Up and Top-Down Attention for Image Captioningand Visual Question Answering. In CVPR.
\bibitem{b10} Ruotian Luo, Brian Price, Scott Cohen and Gregory Shakhnarovich. 2018. Discriminability objective for training descriptive captions. In CVPR.
\bibitem{b11} Kelvin Xu, Jimmy Lei Ba, Ryan Kiros, Kyunghyun Cho, Aaron Courville, Ruslan Salakhutdinov, Richard S. Zemel and Yoshua Bengio. 2015. Show, Attend and Tell: Neural Image Caption Generation with Visual Attention. In ICML.
\bibitem{b12} Peter Anderson, Basura Fernando, Mark Johnson and Stephen Gould. 2016. SPICE: Semantic Propositional Image Caption Evaluation. In ECCV.
\bibitem{b13} Tomas Mikolov, Ilya Sutskever and Kai Chen. 2014. Distributed Representations of Words and Phrases and their Compositionality. In NIPS.
\bibitem{b14} Qi Wu, Chunhua Shen, Lingqiao Liu, Anthony Dick, Anton van den Hengel. 2016. What Value Do Explicit High Level Concepts Have in Vision to Language Problems? In CVPR.
\bibitem{b15} Oriol Vinyals, Alexander Toshev, Samy Bengio, Dumitru Erhan. 2014. Show and Tell: A Neural Image Caption Generator. In CVPR.
\bibitem{b16} Junhua Mao, Wei Xu1 Yi Yang Jiang Wang Alan L. Yuille. 2015. Explain Images with Multimodal Recurrent Neural Networks. In ICLR.
\bibitem{b17} Bo Dai Sanja Fidler Raquel Urtasun Dahua Lin. 2017. Towards Diverse and Natural Image Descriptions via a Conditional GAN  In ICCV.

\bibitem{b18} Zhou Ren1 Xiaoyu Wang1 Ning Zhang1 Xutao Lv1 Li-Jia Li. 2017. Deep Reinforcement Learning-based Image Captioning with Embedding Reward. CVPR.
\bibitem{b19} Jiasen Lu, Caiming Xiong , Devi Parikh , Richard Socher1. 2017. Knowing When to Look: Adaptive Attention via a Visual Sentinel for Image Captioning. In CVPR.
\bibitem{b20} Nannan Li, Zhenzhong Chen. 2018. Image Captioning with Visual-Semantic LSTM. In IJCAI.
\bibitem{b21} Ting Yao, Yingwei Pan, Yehao Li, and Tao Mei. 2018. Incorporating Copying Mechanism in Image Captioning for Learning Novel Object. In CVPR.
\bibitem{b22} Jiasen Lu Jianwei Yang Dhruv Batra Devi Parikh. 2018. Neural Baby Talk. In CVPR.
\bibitem{b23} Di Lu , Spencer Whitehead , Lifu Huang ,Heng Ji , Shih-Fu Chang. 2018. Entity-aware Image Caption Generation. In ACL. Brussels, Belgium Association for Computational Linguistics.
\bibitem{b24} Matt J. Kusner, Yu Sun, Nicholas I. Kolkin, Kilian Q. Weinberger. 2015. From Word Embeddings To Document Distances. International Conference on MachineLearning, Lille, France, 2015.
\bibitem{b25} Girish Kulkarni, Visruth Premraj, Vicente Ordonez, Sagnik Dhar, Siming Li, Alexander C. Berg. 2011. BabyTalk: Understanding and Generating Simple Image Descriptions. In CVPR.
\bibitem{b26} Steven J. Rennie1, Etienne Marcheret1, Youssef Mroueh, Jerret Ross and Vaibhava Goel1. 2017. Self-critical Sequence Training for Image Captioning. In CVPR.
\bibitem{b27} Ryan Kiros, Richard S. Zemel. 2014. Multimodal Neural Language Models. In ICML.
\bibitem{b28} Bo Dai, Dahua Lin. 2017 Contrastive Learning for Image Captioning. In NIPS.
\bibitem{b29} Jiuxiang Gu, Jianfei Cai, Gang Wang, Tsuhan Chen. 2018. Stack-Captioning: Coarse-to-Fine Learning for Image Captioning. In AAAI.
\bibitem{b30} Desmond Elliott and Frank Keller. 2014. Comparing Automatic Evaluation Measures for Image Description. Baltimore, Maryland, USA. Association for Computational Linguistics.

\bibitem{b31} Fuhai Chen, Rongrong Ji, Xiaoshuai Sun, Yongjian Wu, Jinsong Su. 2018. GroupCap: Group-based Image Captioning with Structured Relevance and Diversity Constraints. In CVPR.

\bibitem{b32} Jeff Donahue, Lisa Anne Hendricks, Sergio Guadarrama and Marcus Rohrbach. 2015. Long-term Recurrent Convolutional Networks for Visual Recognition and Description. In CVPR. 
\bibitem{b33} Dzmitry Bahdanau, KyungHyun Cho, Yoshua Bengio. 2015. Neural Machine Translation by Jointly Learning to Align and Translate. In ICLR. 

\bibitem{b34} Jyoti Aneja, Aditya Deshpande, Alexander G. Schwing. 2018. Convolutional Image Captioning. In CVPR
\bibitem{b35} Andrej Karpathy, Li Fei-Fei. 2015. Deep Visual-Semantic Alignments for generating Image Descriptions. In CVPR.
\end{thebibliography}
\end{document}